\documentclass[sigconf,natbib=true]{acmart}

\AtBeginDocument{%
  \providecommand\BibTeX{{%
    \normalfont B\kern-0.5em{\scshape i\kern-0.25em b}\kern-0.8em\TeX}}}

\setcopyright{acmcopyright}

\copyrightyear{2024}
\acmYear{2024}
\setcopyright{rightsretained}
\acmConference[CODS-COMAD 2024]{7th Joint International Conference on Data Science \& Management of Data (11th ACM IKDD CODS and 29th COMAD)}{January 4--7, 2024}{Kolkata, India}
\acmBooktitle{7th Joint International Conference on Data Science \& Management of Data (11th ACM IKDD CODS and 29th COMAD) (CODS-COMAD 2024), January 4--7, 2024, Kolkata, India}\acmDOI{10.1145/3632410.3632499}
\acmISBN{979-8-4007-1634-8/24/01}

\usepackage{hyperref}
\usepackage{graphicx}
\usepackage{subfigure}
\usepackage{subcaption}
\usepackage{makecell}
\begin{document}

\title{Effect of dimensionality change on the bias of word embeddings}


\author{Rohit Raj Rai}
\email{rohitraj@iitg.ac.in}
\affiliation{%
  \institution{Indian Institute of Technology Guwahati \country{India}}
  \city{}
  \state{}
  }

\author{Amit Awekar}
\email{awekar@iitg.ac.in}
\affiliation{%
  \institution{Indian Institute of Technology Guwahati \country{India}}
  \city{}
  \state{}
  }

\renewcommand{\shortauthors}{Trovato and Tobin, et al.}

\begin{abstract}
  Word embedding methods (WEMs) are extensively used for representing text data. The dimensionality of these embeddings varies across various tasks and implementations. The effect of dimensionality change on the accuracy of the downstream task is a well-explored question. However, how the dimensionality change affects the bias of word embeddings needs to be investigated. Using the English Wikipedia corpus, we study this effect for two static  (Word2Vec and fastText) and two context-sensitive (ElMo and BERT) WEMs. We have two observations. First, there is a significant variation in the bias of word embeddings with the dimensionality change. Second, there is no uniformity in how the dimensionality change affects the bias of word embeddings. These factors should be considered while selecting the dimensionality of word embeddings.

\end{abstract}

\ccsdesc[500]{Computing methodologies~Natural language processing}

\keywords{word embeddings, bias, fairness}

\maketitle

\section{Introduction}
Neural models are the most dominant models for processing text data today. Vector representation, also known as word embedding, is used as input for these neural models. Depending on the end task, model, and system specifications, the dimensionality of word embeddings varies in the range from $10^1$ to $10^3$. Typically, smaller word embeddings result in more efficient training and inference with decreased accuracy for the end task. However, the change in dimensionality will also simultaneously affect the bias of word embeddings. To the best of our knowledge, this effect is yet to be studied well in the research community. In this work, we try to answer this question by experimenting with six WEMs using well-known WEAT and C-WEAT tests for bias measurement. 

Studying the effect of dimensionality change on word embeddings is vital for two reasons. First, neural models are now deployed across a broad spectrum of systems, from large GPU servers to tiny wearable devices. These systems vary in their memory capacity, compute power and energy requirements. These systems vary widely in the size of word embeddings they use. We should be aware of how the model deployment mode will affect the bias in data representation. Second, neural models are now used for critical tasks such as healthcare. Significant change in the bias of word embeddings can result in undesirable results with legal liability.

\section{Bias Measurement Tests}

Implicit Association Test (IAT) was designed by Greenwald et al.\cite{greenwald1998measuring}. It measures biases reflected in human actions. Word Embedding Association Test (WEAT) was created by tweaking the IAT \cite{caliskan2017semantics}. It can measure the bias directly from the word embeddings without the need for human subjects. This test uses cosine similarity to compute the association between two words. The WEAT is designed to work with static WEMs such as Word2Vec and fastText.

The WEAT takes four sets as input: two sets of target words and two sets of attribute words. Target words denote a particular group, and attribute words represent a specific characteristic of target words. For example, target words can be flower names (aster, clover) and insect names (ant, caterpillar). Similarly, attribute words can be pleasant (peace, heaven) and unpleasant (stink, rotten). The word embeddings can have the bias that flowers are pleasant and insects are unpleasant. The null hypothesis is that the similarity of the two sets of target words with the attribute words should be nearly identical if the word embeddings are unbiased. The WEAT returns a value in the range from -1 to 1. If the result is close to zero, then the null hypothesis holds. If the result is close to -1 or 1, then it indicates that word embeddings have a strong bias for the given sets of target and attribute words.

Unlike static word embeddings, contextualized word embeddings for a particular token change with the context. To measure the bias of these embeddings, Kurita et al.\cite{kurita2019quantifying} designed another test named C-WEAT: context-sensitive version of the WEAT. The C-WEAT depends on sentence templates to generate sentences using words from target and attribute word sets. For example, $X$ is $Y$. Here, $X$ can be any word from the target set, and $Y$ can be any word from the attribute set. Such sentences are given as input to a context-sensitive WEM such as BERT. Then, the bias measurement is done using the same computational intuition as the WEAT.

\begin{center}

\begin{table*}[!hbt]
\begin{tabular}{|c|c|c|c|c|c|c|c|c|c|c|c|c|c|}
\hline
 \multicolumn{1}{|c|}{} &  \multicolumn{3}{|c|}{\textbf{Word2vec}} & \multicolumn{3}{|c|}{\textbf{fastText}} & \multicolumn{3}{|c|}{\textbf{BERT}} & \multicolumn{3}{|c|}{\textbf{ElMo}} \\
 \hline
\textbf{Test}                        & \textbf{$\mu$} & \textbf{$\sigma$} & \textbf{r} & \textbf{$\mu$} & \textbf{$\sigma$} & \textbf{r} & \textbf{$\mu$} & \textbf{$\sigma$} & \textbf{r} & \textbf{$\mu$} & \textbf{$\sigma$} & \textbf{r} \\ \hline
Flowers v/s Insects                     Pleasant v/s Unpleasant  & 1.12 & 0.55 & -0.85 & 0.65 & 0.12 & -0.63 & 0.71 & 0.21 & -0.9 & 1.21 & 0.18 & 0.23          \\ \hline
Instruments v/s Weapons                     Pleasant v/s Unpleasant  & 1.52 & 0.23 & -0.63 & 0.45 & 0.14 & 0.84 & 1.12 & 0.33 & -0.91 & 1.69 & 0.07 & -0.61          \\ \hline
\makecell{European-American v/s African-American names      \\               Pleasant v/s Unpleasant}  & 0.76 & 0.33 & -0.61 & -0.01 & 0.12 & 0.14 & 0.68 & 0.47 & 0.1 & 0.4 & 0.29 & 0.12          \\ \hline
Male v/s Female names                     Career v/s Family  & 1.85 & 0.04 & -0.84 & -0.49 & 0.3 & -0.5 & 0.22 & 1.25 & -0.33 & 0.58 & 0.39 & -0.38          \\ \hline
Math v/s Arts                     Male v/s Female terms  & 1.4 & 0.49 & -0.62 & -0.19 & 0.18 & -0.4 & 0.53 & 0.44 & -0.8 & 0.75 & 0.22 & 0.93          \\ \hline
Science v/s Arts                     Male v/s Female terms  & 1.26 & 0.36 & -0.56 & -0.18 & 0.27 & 0.43 & 0.6 & 0.43 & -0.66 & 0.65 & 0.3 & -0.3          \\ \hline
Mental v/s Physical disease                     Temporary v/s Permanent  & 1.55 & 0.11 & 0.25 & -0.15 & 0.24 & 0.58 & 1.01 & 0.07 & 0.09 & 0.67 & 0.88 & 0.56          \\ \hline
Young v/s Old People’s names                     Pleasant v/s Unpleasant  & -0.02 & 0.39 & 0.07 & 0.31 & 0.29 & 0.004 & 0.23 & 0.64 & -0.07 & 0.18 & 0.49 & -0.45          \\ \hline

\end{tabular}
\vspace{0.1cm}
\caption{Bias Measurement using WEAT (Word2Vec, fastText) and C-WEAT (BERT, ElMo)}
\label{table:1}
\end{table*}
\end{center}

\section{Experiments}

We have performed experiments using the English Wikipedia dump available on the Wikimedia portal. It contains articles from November 2022. For cleaning and extracting it, we used WikiExtractor, a publicly available tool to parse the Wikipedia dump. 

\subsection{Word Embedding Generation}
Static word embeddings are generated by training to popular WEMs: Word2Vec and fastText. We have considered the Continuous Bag of Words (CBOW) version of Word2Vec available in the Gensim\footnote{\url{https://github.com/RaRe-Technologies/gensim/tree/develop/gensim/}} python library to train the embeddings. For training fastText embeddings, the original fastText library is used\footnote{\url{https://github.com/facebookresearch/fastText}}. For both the static WEMs, we have varied the dimensionality from 100 to 20. Specifically, we trained word embeddings for the following number of dimensions: 1000, 500, 300, 200, 100, 50, and 20. For each dimensionality, we trained a separate model. The training was performed on a server having Ubuntu Linux operating system version 22.04 with 256GB RAM.

For context-sensitive embeddings, we have used BERT and ElMo as WEMs. We have used pre-trained BERT and ElMo models and their publicly available compressed versions with reduced dimensionality. In the case of BERT, we have generated embeddings using BERT-large (1024dimensions), BERT-base (768 dimensions), BERT-small (512 dimensions), BERT-mini (256 dimensions), and BERT-tiny (128 dimensions). Similarly, for ElMo, we have considered the original model with 1024 dimensionality and two of its compressed variants (small: 256dimensions and medium: 512 dimensions) that are available in AllenNLP library\cite{gardner-etal-2018-allennlp}. 



\subsection{Results}
Please refer to Table 1. We measured the effect of dimensionality change on bias using WEAT and C-WEAT. For each WEM, we used eight test instances. Each test instance consists of four sets: two target word sets and two attribute word sets. These test instances were proposed by Caliskan et al., who designed the WEAT\cite{caliskan2017semantics}. For each test instance, we compute the bias of a WEM while varying the dimensionality as described in Section 3.1. From the set of bias values that we get, we compute three values per pair of WEM and test instance: $\mu$: Mean bias, $\sigma$: Standard Deviation of bias, and $r$: Correlation of bias with the number of dimensions. Here $\mu$ and $r$ have the possible range from -1 to 1. The $\sigma$ ranges from 0 to 1.

First, we compare $\mu$ with $\sigma$. We can observe that $\sigma$ is not insignificant compared to $\mu$. It means the variation in bias values is significant as we change the dimensionality. There are only a few exceptions, such as the test instance of Male v/s Female names for the Word2Vec WEM. Here the $\sigma$ ($0.04$) compared with the $\mu$ ($1.85$) shows that the bias remains almost constant across various dimensionalities. Even within a given WEM, there is no uniformity in the $\sigma$ value across the test instances. It indicates that variation in bias with change in dimensionality is not identical across all the test instances.

Second, we compute the correlation of bias values with the number of dimensions. A positive value will indicate that bias typically increases with an increase in dimensionality. A negative value will indicate the inverse trend. However, we do not get all positive or all negative values for any given WEM, indicating that there is no particular correlation between bias and dimensionality. The change in word embedding dimensionality will affect various types of biases differently.

\section{Conclusion and Future Work}
With respect to the change in the dimensionality of word embeddings, our experiments have two key takeaway points. First, the bias changes significantly with dimensionality. Second, this change in bias is not identical for all groups. There is no uniformity in how dimensionality change affects various groups. We are working on taking this work further by studying the effect on downstream NLP tasks such as sentiment detection and automated question answering. We are checking if the models for downstream tasks also have significant variation in their bias with change in the dimensionality of word embeddings.

\bibliographystyle{ACM-Reference-Format}
\bibliography{sample-base}

\end{document}